\documentclass[runningheads]{llncs}
\usepackage{xcolor}

\usepackage[T1]{fontenc}
\usepackage{graphicx}
\usepackage{booktabs}
\newcommand{\cotton}{cotton}

\begin{document}

\title{On the Transferability of Agricultural Weed Detection Under Cross-Field Distribution Shift}
\author{Nikhilesh Prabhakar\inst{1} \and
Pranuthi Tenali\inst{1} \and 
Wilfredo Abudeye Fernandez\inst{2} \and
Shekhar Borah\inst{2} \and
Athresh Karanam\inst{1} \and
Erik Blasch\inst{3} \and
Prabha Sundaravadivel\inst{2} \and
Sriraam Natarajan\inst{1}}
\authorrunning{Prabhakar et al.}
%
\institute{The University of Texas at Dallas \\
\email{\{nikhilesh.prabhakar, pranuthi.tenali, athresh.karanam, sriraam.natarajan\}@utdallas.edu}\and
The University of Texas at Tyler \\
\email{wabudeyefernandez@patriots.uttyler.edu} \\
\email {\{psundaravadivel, sborah\}@uttyler.edu}\and
Air Force Research Lab \\
\email{erik.blasch.1@us.af.mil}}
\maketitle
\begin{abstract}

Accurate agricultural weed detection in real-world field conditions is essential for precision agriculture, enabling targeted intervention and reducing yield loss. Recent work has reported strong detection performance from UAV-based imagery across a range of crops, yet existing approaches evaluate within a single crop and field, leaving practitioners with little evidence that a model trained on one crop will generalize to a new field or crop type. In this work, we characterize where cross-dataset weed-localization performance degrades and which modeling choices recover it, reducing the need to relabel every new deployment field. We introduce a newly collected and annotated UAV image dataset for agricultural weed detection in cotton fields and use it alongside an existing soybean dataset collected under a similar protocol. Using these datasets, we evaluate the performance of several strategies for transferring a detector trained on one crop to another, comparing unsupervised domain adaptive object detection (DAOD) against pretraining on a domain-adjacent source dataset followed by few-shot fine-tuning on the target dataset. Our analysis spans target-domain label budgets from zero to the full target dataset, characterizing the trade-off between adaptation strategy and annotation effort. We find that few-shot fine-tuning with as few as 25 labeled target examples outperforms unsupervised DAOD in our cross-crop comparison, suggesting that source domain selection combined with modest target supervision is more productive than algorithmic sophistication in adaptation.

\end{abstract}

\section{Introduction}
Weeds are a major contributor to yield loss in row crop agriculture, competing with crops for nutrients, water, and light~\cite{oerke2006crop}. Timely and accurate weed detection is therefore central to effective field management and to reducing the indiscriminate use of herbicides, which carries substantial economic and environmental costs~\cite{owen2005herbicide}. Unmanned aerial vehicles (UAVs) have emerged as a practical platform for scaling weed monitoring to entire fields~\cite{torres2013uav,lottes2017uav}. They offer high-resolution imagery at low operational cost and can be deployed with minimal infrastructure relative to ground-based systems.

Modern weed detection pipelines on UAV imagery rely predominantly on deep learning approaches~\cite{bah2018deepUAVdetection,veeranampalayam2020comparison}. The dominant paradigm involves training an object detector on a labeled dataset collected from the field of interest. While this approach can yield strong detection performance, it imposes high annotation costs. Producing high-quality bounding box labels requires expert knowledge of crop and weed species, and the resulting datasets are often narrow in scope, covering a single location, growth stage, or weed species~\cite{hu2024weedDetectionSurvey}. Models trained under these conditions tend to generalize poorly when deployed in new fields, seasons, or to different crops, which limits the practical utility of any single labeled dataset~\cite{borah2025unmanned,hu2024weedDetectionSurvey}.

This challenge is particularly relevant to the Dynamic Data-Driven Applications Systems (DDDAS) paradigm ~\cite{dddas}, in which models guide the collection of data through physical systems that is then incorporated to improve decision-making. In weed detection, deploying such systems is challenging due to differences in field conditions, crop types, and weed population. In this work we study how much target domain data is needed, and how to effectively leverage this new data to aid in developing practical data-driven monitoring systems.

Domain Adaptive Object Detection (DAOD) offers a principled approach to leverage labeled data from a related source domain to improve detection performance on an unlabeled or sparsely labeled target domain. Recent frameworks such as ALDI~\cite{Kay2025ALDI} have standardized benchmarking across DAOD methods and demonstrated strong results on established benchmarks. However, the application of DAOD to UAV-based agricultural imagery in general and weed detection in particular remains underexplored. Existing work in this direction has focused primarily on weather-induced domain shift~\cite{mengment2025TTSDA}, leaving the more common practical scenario of cross-field, cross-crop transfer largely unexamined.

In this work, we address this gap by presenting {\bf an empirical study of cross-crop transfer for UAV-based weed detection.} We introduce a newly collected and annotated UAV image dataset for weed detection in \cotton{} fields and use it alongside a previously published soybean dataset collected under a similar protocol~\cite{borah2025unmanned}. Using these datasets we evaluate a range of strategies for transferring a detector trained on one crop to another, comparing unsupervised DAOD against a simpler alternative: pretraining on a domain-adjacent source dataset and fine-tuning with a small number of labeled examples from the target domain. We, thus, study the data-assimilation aspect of the DDDAS paradigm by evaluating how detection performance changes as the number of available target samples increases. Our central finding is that this latter strategy is remarkably effective, with fine-tuning a detector trained on a domain-adjacent source using as few as $k{=}25$ labeled target examples outperforming DAOD on the same target. A key aspect of our setup is that we examine performance across a spectrum of target domain label budgets, from $k{=}0$ (no examples from target domain) up to the full target dataset, which allows us to characterize the trade-off between adaptation strategy and annotation effort.

We make the following key contributions: (1) We present a newly collected and annotated UAV image dataset for weed detection in \cotton{} fields, providing a real-world benchmark for evaluating cross-crop transfer in agricultural object detection settings. (2) We design and conduct a systematic empirical comparison of unsupervised DAOD against domain-adjacent pretraining with few-shot fine-tuning across a range of target domain label budgets.

Our analysis suggests that in settings where small amounts of target-domain supervision are practical to obtain, such as cross-crop transfer in UAV-based weed detection, {\bf source domain selection combined with few-shot fine-tuning may be more productive in practice than algorithmic sophistication in adaptation.}

\section{Background and Related Work}

\subsection{Weed Detection in UAV Imagery}
UAVs have become an increasingly common platform for weed detection in precision agriculture, offering high-resolution imagery at field scale at a low operational cost. Over the last decade, deep learning approaches have become predominant for processing this imagery~\cite{bah2018deepUAVdetection,veeranampalayam2020comparison}. A variety of architectures have been explored across crops including soybean~\cite{borah2025unmanned} and cotton~\cite{das2026uavCotton}. YOLO-based detectors~\cite{redmon2016YOLO} have emerged as a particularly common choice due to their balance of accuracy and real-time inference~\cite{borah2025unmanned,shahi2023uavWeedComparativeStudy}.

A persistent challenge across this line of work is the visual similarity between weed and crop species from overhead viewpoints. This makes both inter-class separation and model generalization difficult~\cite{bah2018deepUAVdetection}. A direct consequence of this lack of generalization is the aspect of poor transferability of models across fields, seasons, and crop types. Weed datasets are typically collected with limited variability in conditions, often from a single location, growth stage, and weed species, which hinders generalization to new field conditions and different crops~\cite{hu2024weedDetectionSurvey}. Domain adaptation has been identified as a promising yet underexplored direction for weed detection~\cite{hu2024weedDetectionSurvey}. Prior work in this direction has primarily examined ground-level crop-weed imagery~\cite{ilyas2023overcoming}, and its application to UAV-based detection pipelines remains limited. In this work, we aim to address this gap by presenting an empirical analysis of domain adaptive object detection methods on a combination of UAV-acquired weed detection datasets, including a newly collected \cotton{} field dataset.

\subsection{Domain Adaptive Object Detection}
Domain Adaptive Object Detection addresses the performance degradation that arises when a detector trained on a labeled source domain is evaluated on a (potentially unlabeled) target domain with a different data distribution. Existing approaches can be broadly classified into two categories - alignment-based methods and self-distillation or self-training methods.

\noindent\textbf{Alignment-based methods} try to reduce the discrepancy between source and target domains by learning domain-invariant feature representations~\cite{ganin2016dann}. Refinements in this direction include Strong-Weak Distribution Alignment~\cite{saito2019swda}, which argues for weak global alignment while enforcing stronger local alignment on detection-relevant regions, and a suite of methods that further incorporate multi-level~\cite{chen2018dafrcnn}, class-aware~\cite{vs2021megacda}, scale-aware~\cite{chen2021sada}, foreground/background-aware~\cite{he2025differentialDAOD} among others, aiming to align the aspects of the representation most relevant for object recognition rather than the entire feature distribution.

\noindent\textbf{Self-distillation/self-training approaches} based on the mean-teacher paradigm generate pseudo-labels on the target domain to supervise adaptation, with the teacher maintained as an exponential moving average of the student~\cite{deng2021umt,li2022cdAdaptiveTeacher}. Since pseudo-label quality bounds the effectiveness of this paradigm, refinements have largely focused on producing or exploiting cleaner supervision through strong-weak augmentation~\cite{li2022cdAdaptiveTeacher}, uncertainty-aware filtering of unreliable predictions~\cite{chen2022probTeacher}, and contrastive objectives that derive useful signal even from noisy bounding boxes~\cite{cao2023contrastiveMeanTeacher} among others.

\section{Datasets}
Our evaluation uses four datasets of aerial and ground weed imagery that vary in crop, weed species, altitude, and sensor. The \textbf{soybean} dataset~\cite{borah2025unmanned} was collected from a weed-infested soybean field in Stoneville, Mississippi: 637 high-resolution RGB images captured by a UAV at 25\,ft, annotated with bounding boxes for pigweed and tiled into 1628 images of size $640\times640$. We collected a new \textbf{cotton} dataset from weed-infested fields in the same region following the protocol of~\cite{borah2025unmanned}: an IF1200A UAV flown at $13$--$15$\,mph and 25\,ft (ground resolution $\approx$2.13\,cm/pixel) over a $\sim$2.3\,km survey, yielding 590 images manually annotated with LabelImg and tiled into 5706 images of size $640\times640$. The remaining two are public: \textbf{CoFly-WeedDB}~\cite{Krestenitis2022} (low-altitude drone, 207 images) and \textbf{CottonWeedDet12}~\cite{Dang2023} (ground-level cotton-weed, 5648 images, 12 classes remapped to a single weed class). The supplementary material gives the collection and reports the splits and box scale (Table~A2), which span nearly three orders of magnitude and are not equalized by preprocessing, and shows representative tiles (Figure~A1).

\section{Methodology}

Weed detection requires carefully collecting and manually annotating large image datasets which is labor-intensive. Consequently, deploying a pretrained detector on the target field is highly desirable. However, the source and target fields are often very different. A key question in cross-field weed detection is how much target-field supervision is required to achieve satisfactory performance. Formally, given pretrained detector on the source domain, and $n$ examples from the target training set, of which $k$ are labeled, the task is to adapt the source detector on the target domain.
To systematically study this, we analyze the following techniques on the transferability among the $4$ different domains.

\subsection{Domain Adaptative Object Detection }
We compare the two dominant families of domain-adaptive object detection.
\\
\textbf{\emph{Adversarial feature alignment} (Align)} adds image-level and instance-level domain classifiers trained through a gradient-reversal layer, encouraging domain-invariant features, without requiring a teacher. \\
\textbf{\emph{Self-distillation} (Distill)} instead maintains an exponential moving average (EMA) copy of the detector as a teacher that produces soft targets on the unlabeled target domain, combined with masked-image consistency (MIC). It doesn't use adversarial alignment. In our evaluation, gradient clipping (max norm $1.0$) is applied to the distillation runs for numerical stability.

We use ALDI (Align-and-Distill)~\cite{Kay2025ALDI}, a unified framework for domain-adaptive object detection, to run our adaptation experiments. ALDI implements both adaptation families above on a common Faster R-CNN backbone and under a single training recipe, standardizing the components that otherwise confound cross-method comparison: strong augmentation, an EMA teacher, and training from a source-only checkpoint. Adversarial alignment and self-distillation therefore differ only in their adaptation objective, which lets us attribute differences in target performance to the adaptation mechanism rather than to incidental training choices. We use this shared recipe for every adaptation run reported below.

\subsection{Few-Shot Fine Tuning}
As a supervised alternative for adaptation, we fine-tune the source-only detector on a small number of labeled \emph{target} tiles. We vary the budget $K \in \{5, 10, 25\}$ labeled images and, for each $K$, draw $M{=}5$ independent random subsets to capture the variance from which images are labeled.
Fine-tuning is standard supervised training, with out any adaptation losses or unlabeled target data, for $1.5$k iterations at learning rate $0.005$, starting from the source checkpoint as pretrained weights. 
\section{Experiments}
In our empirical evaluation, we aim to answer the following questions:

\noindent\textbf{Q1.} Does a detection model trained on one field effectively transfer to another?\\
\textbf{Q2.} Does unsupervised domain adaptation help improve the transfer?\\
\textbf{Q3.} When domain adaptation falls short, does few-shot learning work better?
\subsection{Experimental Setup}
\noindent\textbf{Training:} All detectors are Faster R-CNN with a ResNet-50-FPN backbone, initialized from COCO pretraining, inside the ALDI framework~\cite{Kay2025ALDI}. We use ALDI's fair-baseline recipe throughout: strong augmentation, random erasing, and an exponential moving average (EMA) of the weights for evaluation. Source-only models train for $8$k iterations (SGD, batch size $8$, base learning rate $0.01$, step decay at $7$k). Every domain adaptation run resumes from the source-only checkpoint and trains for a further $10$k iterations on labeled source plus unlabeled target. All datasets reach the detector at a common $640$\,px input: the UAV surveys are tiled, and the others are resized to a $640$\,px shorter edge.

\noindent\textbf{Input Normalization:} All datasets are presented to the detector at a common $640$\,px input. The high-resolution UAV surveys (cotton, soybean; $\approx$ $5184\times3888$\,px), in which weeds are small relative to the frame, are tiled to non-overlapping $640\times640$\,px tiles; the remaining datasets are
resized to a $640$\,px shorter edge.

\noindent\textbf{Evaluation:} All numbers are mAP@50 on the held-out target \emph{test} split. For each run, we select the checkpoint with the best target-validation mAP@50 and report its target-test score. The source-only baselines are instead selected on the source validation split. Domain-adaptation and few-shot results are reported as mean\,$\pm$\, standard deviation over multiple training seeds (and, for few-shot, over random label draws). The training seed varies data ordering, augmentation sampling, and RPN/ROI mini-batch composition. The train/val/test partition is held fixed across all runs.

    \begin{table}[t]
    \centering
    \caption{Domain-adaptation comparison for cross-field weed detection (target-test mAP$_{50}$, checkpoints selected on the target validation split). \emph{In-Domain} is the target's in-domain supervised mAP$_{50}$; \emph{Source-only} is the source-trained detector applied zero-shot to the target. Distillation and adversarial alignment are reported as mean\,$\pm$\,std over $n$ seeds (single-seed cells show no interval); $\Delta$ is the change relative to source-only. Adversarial alignment on the $108$-image CoFly source diverges (Inf/NaN) at the
    default learning rate and is omitted ($\dagger$). Adversarial alignment was not run for the CottonWeedDet12 target (\texttt{---}); we report distillation only.}

    \label{tab:da-comparison}
\begin{tabular}{l c c c c }  
    \toprule
    Source $\rightarrow$ Target & In-Domain & Source-only & Distill ($n$) & Align ($n$) \\
    \midrule
    \multicolumn{5}{l}{\emph{Target: soybean}} \\
    cotton $\rightarrow$ soybean       & 76.8 & 57.8 & $50.3 \pm 2.5$ (5) & $42.8 \pm 4.5$ (5) \\
    cofly $\rightarrow$ soybean        & 76.8 & 4.1  & $\mathbf{16.7 \pm 2.1}$ (3) & div.$^{\dagger}$  \\
    cwdet12 $\rightarrow$ soybean      & 76.8 & 6.0  & $9.6 \pm 1.0$ (3)  & 6.5 (1) \\
    \midrule
    \multicolumn{5}{l}{\emph{Target: cotton}} \\
    soybean $\rightarrow$ cotton       & 37.4 & 7.5  & $8.2 \pm 0.4$ (4)  & $7.4 \pm 0.6$ (5) \\
    cofly $\rightarrow$ cotton         & 37.4 & 1.3  & 1.6 (1)            & 0.7 (1) \\
    cwdet12 $\rightarrow$ cotton       & 37.4 & 1.5  & 3.9 (1)            & 2.5 (1) \\
    \midrule
    \multicolumn{5}{l}{\emph{Target: CottonWeedDet12}} \\
    
    cotton $\rightarrow$ cwdet12       & 94.5 & 31.8 & $\mathbf{42.3 \pm 0.2}$ (3) & --- \\
    soybean $\rightarrow$ cwdet12      & 94.5 & 18.8 & $\mathbf{36.9 \pm 0.5}$ (3) & --- \\
    cofly $\rightarrow$ cwdet12        & 94.5 & 14.3 & $6.3 \pm 0.8$ (3)    & --- \\
    \bottomrule
  \end{tabular}
\end{table}

 \begin{table}[t]
    \centering
    \caption{Few-shot fine-tuning vs.\ unsupervised domain adaptation (target-test
    mAP$_{50}$). Each $K$ column is the mean\,$\pm$\,std over $M$ random label draws
    ($M{=}5$ for $K\!\in\!\{5,10\}$, $M{=}4$ for $K{=}25$). The ``All (full FT)''
    column fine-tunes the source detector on the \emph{entire} target training set
    (full supervision, $n{=}3$ seeds, longer schedule) and is compared against the
    in-domain score to isolate the value of source pretraining.
    \emph{Source} and \emph{DAOD} (from Table~\ref{tab:da-comparison}) and the in-domain score are shown for reference. Fine-tuning on a handful of target labels exceeds both source-only and the best adapted model on the learnable target (soybean (S)), reaching the in-domain score by $K{=}25$, but does not move the target (cotton (C)). CoFly and CWDet12 are referred by CF and CW for brevity.} 
    \label{tab:fewshot}
    \begin{tabular}{l c c c c c c c}
    \toprule
    Transfer & Source & DAOD & $K{=}5$ & $K{=}10$ & $K{=}25$ & Full FT & In-Domain  \\
    \midrule
    C $\rightarrow$ S & 57.8 & 50.3 & $66.7 \pm 4.3$ & $70.8 \pm 3.6$ & $80.0 \pm 3.2$ & $84.9 \pm 0.5$ (3) & 76.8 \\
    S $\rightarrow$ C & 7.5  & 8.3  & $8.8 \pm 1.1$ & $8.2 \pm 0.7$ & $9.4 \pm 1.5$ & $35.1 \pm 0.7$ (3) & 37.4 \\
    CW $\rightarrow$ S & 6.0  & 9.6  & $40.9 \pm 11.4$ & $56.8 \pm 8.2$ & $65.4 \pm 6.2$ & $84.9 \pm 1.6$ (5) & 76.8 \\
    CW $\rightarrow$ C & 1.5  & 3.9  & $2.4 \pm 0.4$ & $3.2 \pm 1.1$ & $3.3 \pm 0.4$ & $43.2 \pm 2.4$ (5) & 37.4 \\
    CF $\rightarrow$ S & 4.1 & 16.7 & $37.2 \pm 15.3$ & $49.5 \pm 13.9$ & $62.9 \pm 2.6$ & $75.4 \pm 1.0$ (3) & 76.8 \\                                
    CF $\rightarrow$ C & 1.3 & 1.6 & $3.0 \pm 1.1$ & $4.1 \pm 1.1$ & $8.1 \pm 2.4$ & $41.3 \pm 2.5$ (3) & 37.4 \\   
    \bottomrule
    \end{tabular}
    \end{table}

\subsection{Results}
We organize the results around the three questions posed above. We first report the quantitative results and then identify the conditions under which it holds and the factors that predict when it breaks.

\noindent\textbf{Q1: Cross-field transfer} 
To answer \textbf{Q1}, by default, a weed detector trained on one source does not always effectively transfer across fields, and its in-domain accuracy is not an indication of how far it will fall. (Table~A1) reports the mAP@50 results for every direct model transfer for every source-to-target pair. For a couple of datasets, Soybean and CWDet12, the in-domain accuracy is strong. The numbers off the diagonal show the transfer mAP@50 scores with no supervised fine-tuning. Most zero-shot transfer collapses to single-digits. The exceptions to this are the transfer between Cotton $\rightarrow$ Soybean (57.8) and Cotton $\rightarrow$ CWDet12 (31.8), the targets that share the species of weed or scale with the source. Pairs that differ in weed-type and scale of detection drop to near zero. 

\noindent\textbf{Q2: Unsupervised domain adaptation.} 
Table~\ref{tab:da-comparison} reports $mAP@50$ scores for source-only, Distill, and Align across our source-target pairs. The results are highly variable. DAOD substantially improves over source-only in some cases (cotton $\rightarrow$ CWDet12; CoFly $\rightarrow$ soybean) but reduces performance in others, including the cotton $\rightarrow$ soybean comparison most relevant to our setting, where Distill and Align fall to 50.3 and 42.8 from a source-only baseline of 57.8. Even where DAOD helps, results often remain far below in-domain training. We note that ALDI was applied with its default configuration and a more extensive hyperparameter or augmentation search may improve results. Our observation is narrower: \emph{applying a modern DAOD framework out of the box does not yield consistent improvements across the pairs we evaluate.} This variability matters in practice. UAV-based weed detection systems are often deployed by agronomists rather than ML specialists, and a method whose success depends on deployment-specific tuning is harder to recommend as a default. Few-shot fine-tuning on a detector pretrained on a domain-adjacent source is a better-understood procedure with fewer tunable modules and less opaque failure modes. We examine whether it offers a more reliable alternative next.

\noindent\textbf{Q3: Few-shot fine-tuning}
Table~\ref{tab:fewshot} presents results for fine-tuning a detector trained on the source domain using a small number of labeled samples from the target domain. A handful of labeled data outperforms DAOD when the target is learnable. On soybean, $k{=}5$ already reaches 66.7±4.3, above both source-only (57.8) and the best DAOD result (50.3), and $k{=}25$ reaches 80.0±3.2, near the 76.8 in-domain score; full fine-tuning reaches 84.9±0.5, indicating that source pretraining adds value. The effect is larger for weak sources: CWDet12 is a weak source for Soybean ($6.0$ for source-only and $9.6$ for best DAOD), yet five labels lift it to a mAP of $40.9$. Harder datasets like cotton behave differently. A handful of labels do not effectively transfer, similar to DAOD.
\section{Conclusion}
We presented an empirical study of cross-crop transfer for UAV-based weed detection, introducing a newly collected cotton dataset and comparing unsupervised DAOD against few-shot fine-tuning on a domain-adjacent source across a range of target label budgets. Applying a modern DAOD framework (ALDI) out of the box yielded inconsistent improvements over source-only training, while few-shot fine-tuning with as few as $k{=}25$ labeled target examples proved more reliable in our setting. This work, thus, provides an initial investigation on the data-assimilation component of the DDDAS paradigm, by examining how limited target-domain information can be incorporated to adapt the detection model to a new domain. Future work will extend this to a full DDDAS loop by incorporating the feedback from the detections inorder to determine the parts of the field to concentrate on, plan optimal flight paths, and design effective monitoring strategies. Extending this analysis to additional crops, a deeper analysis of domain to domain discrepancy in transferability, and leveraging these observations to develop new DAOD methods are promising directions for future work.

\section{Acknowledgements}
The authors gratefully acknowledge the generous support from AFOSR award FA9550-23-1-0239.
\bibliographystyle{splncs04}
\bibliography{refs}
\end{document}


\title{Supplementary Material:\\On the Transferability of Weed Detection\\Under Cross-Field Distribution Shift}
\author{}
\authorrunning{}
\institute{}
\maketitle
\section*{Appendix}

\section{Cross-Field Transfer Matrix}
\label{app:transfer}
\begin{table}[h]
\centering
\caption{Cross-dataset transfer matrix. Each cell reports zero-shot mAP@50 for a model trained on the source (row) and evaluated on the target (column). In-domain results lie on the diagonal (\textbf{bold}). Single seed ($n=1$).}
\label{app:tab:transfer}
\begin{tabular}{lcccc}
\toprule
& \multicolumn{4}{c}{\textbf{Target}} \\
\cmidrule(lr){2-5}
\textbf{Source} & Cotton & Soybean & CoFly & CWDet12 \\
\midrule
Cotton       & \textbf{37.4} & 57.8 & 0.0 & 31.8 \\
Soybean      & 7.5 & \textbf{76.8} & 0.2 & 18.8 \\
CoFly        & 1.3 & 4.1 & \textbf{17.6} & 14.3 \\
CWDet12      & 1.5 & 6.0 & 0.1 & \textbf{94.5} \\
\bottomrule
\end{tabular}
\end{table}

\section{Dataset Details}

\textbf{Soybean.} The soybean dataset~\cite{borah2025unmanned} was collected over a
weed-infested field in Stoneville, Mississippi. It has 637 high-resolution RGB frames flown
by UAV at 25,ft and hand-annotated with pigweed bounding boxes; tiling to $640\times640$
yields 1{,}628 images.

\textbf{Cotton.} We collected the cotton dataset over weed-infested fields in the same
region, following the protocol of~\cite{borah2025unmanned}. Imagery was captured with an
IF1200A UAV chosen for payload, endurance, and field reliability. The flight plan covered
roughly 2.3,km at 13--15,mph and a constant 25,ft altitude, giving a ground resolution near
2.13,cm/px. We annotated weeds by hand with LabelImg. The survey produced 590 frames,
tiled to $640\times640$ into 5{,}706 images.

\textbf{Public datasets.} CoFly-WeedDB~\cite{Krestenitis2022} is a low-altitude drone
dataset of 207 images. CottonWeedDet12~\cite{Dang2023} is a medium-scale ground-level
cotton-field dataset of 5{,}648 images with 12 weed classes, which we remap to the single \texttt{weed} class used throughout.
\label{app:data}
  \begin{table}[h]
  \centering
  \caption{Dataset splits (images / weed boxes) and weed-box scale (median box area
  as a fraction of the image). All datasets use a single \texttt{weed} class. Box
  scale spans nearly three orders of magnitude and is not equalized by preprocessing.}
  \label{tab:datasets}
  \begin{tabular}{lccccc}
  \toprule
  Dataset & Train & Val & Test & Box (\% img) \\
  \midrule
  Cotton (UAV, tiled)   & 4248 / 1004 & 734 / 237  & 724 / 133 & 25.2 \\
  Soybean (UAV, tiled)  & 1156 / 293  & 308 / 77   & 164 / 42  & 25.3 \\
  CoFly-WeedDB          & 108 / 194   & 26 / 43    & 67 / 140  & 4.4 \\
  CottonWeedDet12       & 3966 / 6621 & 841 / 1320 & 841 / 1447 & 10.7 \\
  \bottomrule
  \end{tabular}
  \end{table}

\begin{figure}[h]
\centering
\includegraphics[width=\linewidth]{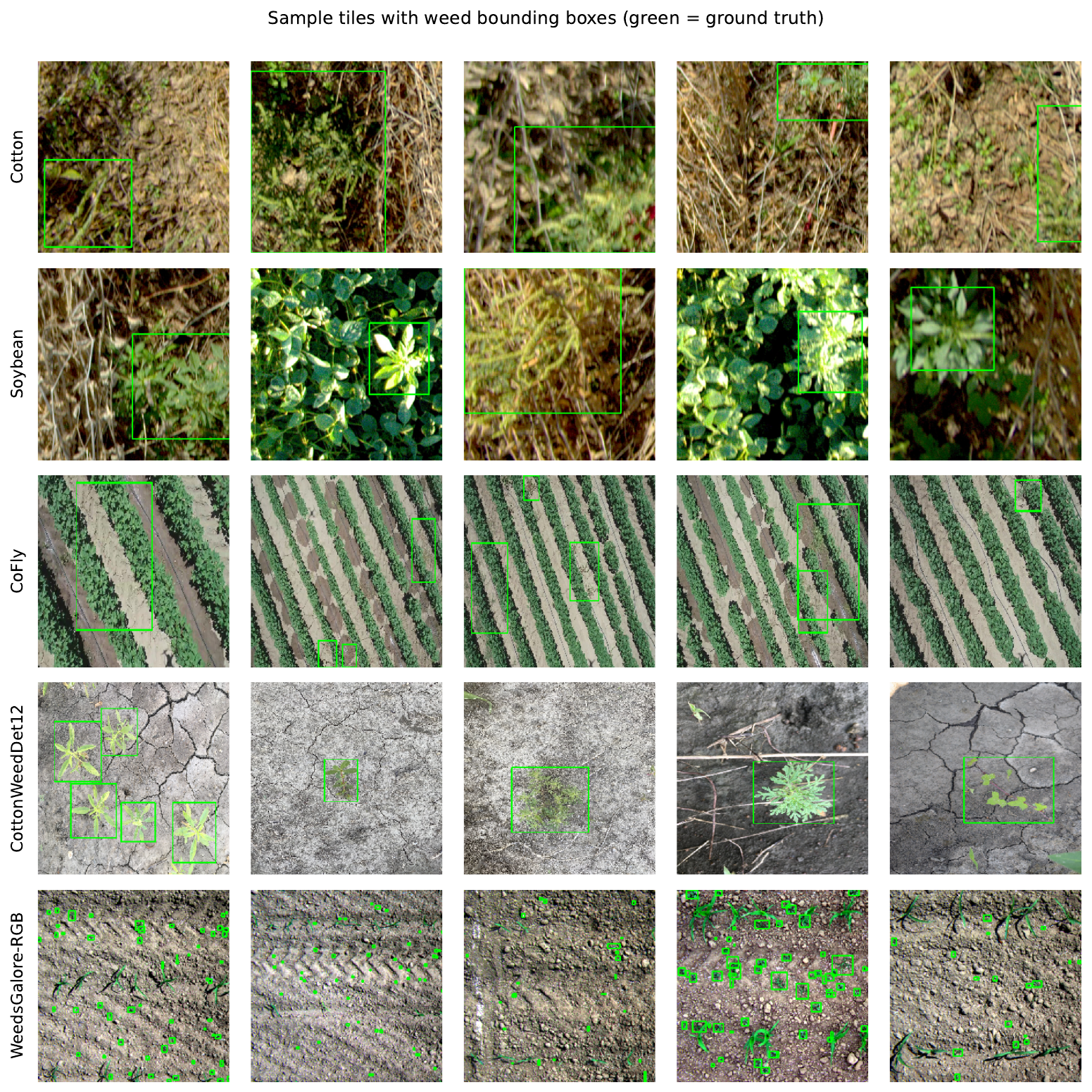}
\caption{Representative tiles from the four datasets. The two UAV surveys (cotton,
soybean) share row structure, soil, and canopy color and are hard to tell apart by
eye; the public sets differ in altitude, viewpoint, and instance scale.}
\label{fig:samples}
\end{figure}
\FloatBarrier
\bibliographystyle{splncs04}
\bibliography{refs.bib}